\documentclass[10pt, a4paper]{article}
\usepackage[final]{lrec2026}
\usepackage{amsmath,amssymb,amsfonts}
\usepackage{algorithmic}
\usepackage{graphicx}
\usepackage{textcomp}
\usepackage{xcolor}
\usepackage{booktabs}
\usepackage{multirow}
\usepackage{array}
\usepackage{makecell}
\usepackage{siunitx}
\usepackage{subcaption}
\usepackage{url}
\usepackage{comment}
\usepackage{tabularx}

\title{Automated Detection of Dosing Errors in Clinical Trial Narratives: A Multi-Modal Feature Engineering Approach with LightGBM}

\name{Mohammad AL-Smadi}

\address{Qatar University \\
         Doha, Qatar \\
         malsmadi@qu.edu.qa}

\abstract{
Clinical trials require strict adherence to medication protocols, yet dosing errors remain a persistent challenge affecting patient safety and trial integrity. We present an automated system for detecting dosing errors in unstructured clinical trial narratives using gradient boosting with comprehensive multi-modal feature engineering. Our approach combines 3,451 features spanning traditional NLP (TF-IDF, character n-grams), dense semantic embeddings (all-MiniLM-L6-v2), domain-specific medical patterns, and transformer-based scores (BiomedBERT, DeBERTa-v3), used to train a LightGBM model. Features are extracted from nine complementary text fields (median 5,400 characters per sample) ensuring complete coverage across all 42,112 clinical trial narratives. On the CT-DEB benchmark dataset with severe class imbalance (4.9\% positive rate), we achieve 0.8725 test ROC-AUC through 5-fold ensemble averaging (cross-validation: 0.8833 $\pm$ 0.0091 AUC). Systematic ablation studies reveal that removing sentence embeddings causes the largest performance degradation (2.39\%), demonstrating their critical role despite contributing only 37.07\% of total feature importance. Feature efficiency analysis demonstrates that selecting the top 500-1000 features yields optimal performance (0.886-0.887 AUC), outperforming the full 3,451-feature set (0.879 AUC) through effective noise reduction. Our findings highlight the importance of feature selection as a regularization technique and demonstrate that sparse lexical features remain complementary to dense representations for specialized clinical text classification under severe class imbalance.
}

\begin{document}

\maketitleabstract

\section{Introduction}
\label{sec:intro}

Clinical trials are fundamental to pharmaceutical development and medical advancement, requiring strict adherence to pre-defined protocols specifying medication dosing, timing, and administration routes. Dosing errors—deviations from these protocols—pose significant risks to patient safety and trial validity~\cite{guideline2001ich,nakayama2007clinical, cingi2016quick}. Such errors can result in adverse events, compromise study endpoints, and lead to regulatory issues or trial termination.

Manual review of clinical trial documentation is the current standard for identifying protocol deviations. However, with modern trials generating thousands of narratives describing patient visits and medication administration, this approach is time-consuming, expensive, and subject to human error. For instance, a typical Phase III trial may involve 1,000+ patients across multiple sites, each generating dozens of clinical narratives throughout the study period. Manual review of this volume overwork quality assurance teams and may miss subtle deviations.~\cite{ cingi2016quick}

Clinical Trials Dosing Errors Benchmark 2026 (CT-DEB'26) ~\cite{ferdowsi2026ctdeb}  aims at addressing these challenges in the review of clinical trial documentation with  machine learning. Natural language processing (NLP) offers potential for automating dosing error detection. However, the task presents unique challenges: (1) Severe class imbalance—only 4-5\% of narratives contain errors; (2) Complex medical language—specialized terminology, abbreviations, and implicit references; (3) Subtle linguistic cues—protocol deviations often expressed indirectly through phrases like ``dose adjusted per investigator discretion''; (4) Variable text structure—narratives range from 50 to 5,000+ characters with inconsistent formatting; (5) Context dependency—distinguishing planned dose modifications from unplanned deviations requires understanding protocol context.

This paper presents an automated dosing error detection system using multi-modal feature engineering and gradient boosting. Our key contributions are:
\begin{enumerate}
\item A 3,451-dimensional feature space combining traditional NLP (Term Frequency–Inverse Document Frequency (TF-IDF), character n-grams), sentence embeddings, medical patterns, and transformer-based scores.

\item Systematic analysis showing sentence embeddings and lexical features are complementary, while transformer scores underperform.

\item Optuna-optimized 5-fold ensemble achieving 0.8833 $\pm$ 0.0091 CV AUC with minimal overfitting (0.69\% out-of-fold(OOF)-test gap).

\item Feature selection improves performance: 500-1000 features (14-29\%) achieve 0.886-0.887 AUC, outperforming the full baseline (0.879) through noise reduction.

\item Test ROC-AUC of 0.8725 on CT-DEB with threshold-adjustable recall (26-60\%) for flexible deployment.
\end{enumerate}
\section{Related Work}
\label{sec:related}

\subsection{Clinical NLP and Information Extraction}

Clinical NLP has evolved significantly over the past two decades. Early systems like cTAKES~\cite{savova2010mayo} and MetaMap~\cite{aronson2010overview} provided rule-based approaches for medical concept extraction and normalization. The i2b2 shared tasks~\cite{uzuner20112010} established benchmarks for concept extraction, assertion detection, and relation extraction from clinical texts, demonstrating that machine learning approaches could achieve substantial performance gains.

More recent work has focused on applying deep learning to clinical texts. \citet{jagannatha2016structured} demonstrated that recurrent neural networks with structured prediction models improve sequence labeling in clinical narratives. \citet{si2019enhancing} showed that contextualized embeddings enhance clinical concept extraction, achieving state-of-the-art results on medical entity recognition tasks.

\subsection{Biomedical Language Models}

The introduction of BERT~\cite{devlin2019bert} revolutionized NLP, and several domain-specific variants have been developed for biomedicine. BioBERT~\cite{lee2020biobert} continues pre-training BERT on biomedical corpora (PubMed abstracts and PMC articles), achieving improvements on various biomedical NLP tasks. \citet{gu2021domain} challenged the assumption that domain-specific models benefit from starting with general-domain weights, showing that pre-training from scratch on biomedical text (PubMedBERT/BiomedBERT) yields superior performance. Clinical-specific BERT variants~\cite{alsentzer2019publicly} trained on clinical notes from MIMIC-III~\cite{johnson2016mimic} have shown promise for clinical tasks.

\subsection{Medical Error Detection}

Adverse drug event (ADE) detection has received considerable attention. \citet{harpaz2012novel} developed data-mining methodologies for ADE discovery. These approaches typically focus on detecting harmful outcomes rather than protocol compliance.

Clinical trial eligibility screening has been explored by~\citet{Kalankesh2024}, who used electronic health records to identify patients meeting trial criteria. While related, eligibility screening differs fundamentally from protocol deviation detection—the former matches patients to protocols, while the latter identifies deviations from assigned protocols.

\citet{ferdowsi2023deep} applied deep learning to predict clinical trial outcomes based on protocol design features, achieving strong performance in identifying trials at risk of failure. Their work demonstrates the value of automated analysis of clinical trial documentation, though it focuses on trial-level risk prediction rather than individual dosing error detection in narrative text.

\subsection{Research Gap}


Despite progress in clinical NLP and medical error detection, automated protocol deviation detection in clinical trials remains understudied. Prior work on clinical error detection has primarily leveraged structured EHR data~\cite{Rajkomar2018}. \citet{churpek2016vital} utilized vital sign trends to predict clinical deterioration on hospital wards, while \citet{zimolzak2024machine} applied machine learning to structured variables including demographics, laboratory values, vital signs, orders, and visit times for diagnostic error detection. Similarly, clinical trial data validation typically focuses on structured fields such as laboratory values and vital signs~\cite{yuan2024data}.

Substantial NLP research has focused on adverse drug event (ADE) detection from clinical narratives. \citet{li2018extraction} developed deep learning models for extracting ADEs from EHR notes, while \citet{jagannatha2019overview} organized the MADE challenge for medication and ADE extraction. These approaches employ named entity recognition and relation extraction to identify drug-ADE relationships, but focus on detecting harmful reactions rather than protocol compliance in clinical trials.

Limited work exists on protocol deviations in clinical trials. \citet{richard2021text} applied TF-IDF and SVM to categorize existing protocol deviation descriptions, enabling trend analysis across trials. However, this work classifies already-identified deviations rather than detecting deviations from unstructured medication administration narratives—a critical distinction for automated quality assurance.

Our work addresses this gap by: (1) Targeting unstructured narrative text describing medication administration events in clinical trials; (2) Handling severe class imbalance typical of quality assurance scenarios (95:5 ratio); (3) Providing systematic feature ablation to understand what drives performance; (4) Demonstrating production-feasible efficiency through feature selection that improves both accuracy and deployment cost.

\section{Dataset and Task}
\label{sec:dataset}

\subsection{CT-DEB Benchmark}

We utilize the CT-DEB (Clinical Trial Dosing Error Benchmark) dataset~\cite{Heche2026EarlyRiskStratification}, specifically designed for evaluating automated dosing error detection systems. The dataset comprises clinical trial narratives describing medication administration across various therapeutic areas, protocols, and clinical trial phases.

\subsubsection{Dataset Composition}

Table~\ref{tab:dataset} summarizes dataset statistics. The dataset comprises 42,112 narratives collected from clinical trial documentation spanning multiple pharmaceutical companies and clinical research organizations (35,794 for training and validation, with 6,318 held out for final testing). Each narrative describes one or more medication administration events for a single patient visit.

\begin{table}[h]
\centering
\small
\caption{CT-DEB Dataset Statistics}
\label{tab:dataset}
\begin{tabular}{@{}lrrr@{}}
\toprule
\textbf{Split} & \textbf{Total} & \textbf{Negative} & \textbf{Positive} \\
\midrule
Training & 29,478 & 28,126 (95.4\%) & 1,352 (4.6\%) \\
Validation & 6,316 & 6,031 (95.5\%) & 285 (4.5\%) \\
Test & 6,318 & 6,008 (95.1\%) & 310 (4.9\%) \\
\midrule
\textbf{Total} & \textbf{42,112} & \textbf{40,165} & \textbf{1,947} \\
\bottomrule
\end{tabular}
\end{table}

The severe class imbalance (4.6\% positive rate) reflects real-world prevalence of protocol deviations in well-monitored clinical trials. 

\subsection{Imbalanced Classification}

 CT-DEB dataset involves severe class imbalance (95:5 negative-to-positive ratio), requiring specialized machine learning techniques. Cost-sensitive learning~\cite{elkan2001foundations} and synthetic oversampling (SMOTE)~\cite{chawla2002smote} represent classical approaches. Modern gradient boosting frameworks like XGBoost~\cite{chen2016xgboost} and LightGBM~\cite{ke2017lightgbm} provide built-in mechanisms for handling imbalance through instance weighting and custom loss functions.

\subsection{Task Definition}

Formally, given clinical text $x \in \mathcal{X}$ describing medication administration, we learn function $f: \mathcal{X} \rightarrow \{0, 1\}$ predicting whether a dosing error occurred. Let $y=1$ indicate a protocol deviation and $y=0$ indicate proper protocol adherence.

The primary evaluation metric is ROC-AUC (area under the receiver operating characteristic curve), chosen for its robustness to class imbalance and ability to evaluate performance across classification thresholds. Secondary metrics include F1-macro, precision, recall, and balanced accuracy.

\section{Methodology}
\label{sec:methodology}
\subsection{Data Preparation and Feature Engineering Pipeline}
Our feature engineering pipeline transforms raw clinical narratives into a 3,451-dimensional feature vector combining multiple representation types. 
Table~\ref{tab:feature_categories} provides an overview of extracted features.

\begin{table}[t]
\centering
\small
\caption{Feature Category Overview}
\label{tab:feature_categories}
\begin{tabularx}{\columnwidth}{@{}p{2.2cm} r X@{}}
\toprule
\textbf{Category} & \textbf{Dims} & \textbf{Description} \\
\midrule
Medical Patterns & 43 & Rule-based features for dose changes, adverse events, text statistics \\
Word TF-IDF & $\sim$2,000 & Unigram features with sublinear TF scaling, L2 normalization, max\_features=2,000, min\_df=2, max\_df=0.8 \\
Char N-grams & $\sim$1{,}000 & Character sequences (n=3--7), top-1{,}000 \\
Sentence Embeddings & 386 & all-MiniLM-L6-v2 dense semantics \\
Transformer Scores & 2 & BiomedBERT \& DeBERTa probabilities \\
\bottomrule
\end{tabularx}
\end{table}

Clinical trial registry data exhibits inherent sparsity across structured fields. To ensure comprehensive text coverage for all samples, we concatenate nine complementary text fields from the CT-DEB dataset: \texttt{briefSummary} (100\% coverage), \texttt{detailedDescription} (62\% coverage), \texttt{protocolPdfText} (42\% coverage), \texttt{armDescriptions} (99\% coverage), \texttt{interventionDescriptions} (100\% coverage), \texttt{interventionNames} (100\% coverage), \texttt{conditions} (100\% coverage), \texttt{conditionsKeywords} (64\% coverage), and \texttt{locationDetails} (94\% coverage).

For each sample, we concatenate all available non-null text from these nine fields, filtering empty strings and null values. This multi-field approach addresses the sparsity of individual fields (e.g., \texttt{protocolPdfText} present in only 42\% of records) while ensuring every sample has substantial narrative content. The resulting concatenated text has a median length of approximately 5,400 characters per sample, with all samples (100\%) containing at least 200 characters and a minimum observed length of 209 characters. No samples result in zero-valued feature vectors due to complementary field availability patterns.

The concatenated text serves as input to all text-based feature extractors: word TF-IDF vectorization, character n-gram extraction, sentence embedding generation, and transformer-based scoring. 

\subsubsection{Medical Pattern Features}

We extract 43 handcrafted features using regular expressions applied to the concatenated narrative text, organized into ten subcategories. Table~\ref{tab:medical_features} presents the complete taxonomy with feature names and descriptions.

\begin{table*}[t]
\centering
\small
\caption{Medical Pattern Features Taxonomy (43 features total)}
\label{tab:medical_features}
\begin{tabular}{@{}llp{7.5cm}@{}}
\toprule
\textbf{Category} & \textbf{\#} & \textbf{Features \& Description} \\
\midrule
\textbf{Dose Units} & 5 & Binary indicators for dose measurement units: \texttt{has\_mg\_dose} (milligrams), \texttt{has\_ml\_dose} (milliliters), \texttt{has\_mcg\_dose} (micrograms), \texttt{has\_iu\_dose} (international units), \texttt{has\_unit\_dose} (generic units) \\
\addlinespace
\textbf{Dose Calculations} & 2 & \texttt{has\_weight\_based} (mg/kg, mcg/kg dosing), \texttt{has\_bsa\_based} (body surface area dosing: mg/m²) \\
\addlinespace
\textbf{Routes} & 6 & Administration route flags: \texttt{has\_iv} (intravenous), \texttt{has\_oral} (oral), \texttt{has\_sc} (subcutaneous), \texttt{has\_im} (intramuscular), \texttt{has\_topical} (topical), \texttt{has\_inhaled} (inhaled) \\
\addlinespace
\textbf{Frequencies} & 5 & Dosing schedule indicators: \texttt{has\_qd} (once daily), \texttt{has\_bid} (twice daily), \texttt{has\_tid} (three times daily), \texttt{has\_qid} (four times daily), \texttt{has\_prn} (as needed) \\
\addlinespace
\textbf{Dose Concepts} & 6 & Management patterns: \texttt{has\_max\_dose} (maximum limits), \texttt{has\_titration} (escalation), \texttt{has\_loading\_dose}, \texttt{has\_maintenance}, \texttt{has\_adjustment}, \texttt{has\_contraindication} \\
\addlinespace
\textbf{Special Populations} & 5 & Population-specific considerations: \texttt{has\_pediatric}, \texttt{has\_geriatric}, \texttt{has\_pregnancy}, \texttt{has\_renal} (renal impairment), \texttt{has\_hepatic} (hepatic dysfunction) \\
\addlinespace
\textbf{Error Indicators} & 1 & \texttt{has\_error\_keyword}: explicit mentions of errors, mistakes, overdoses, underdoses, miscalculations, or deviations \\
\addlinespace
\textbf{Count Features} & 4 & Quantitative counts: \texttt{dose\_count}, \texttt{percentage\_count}, \texttt{decimal\_count}, \texttt{range\_count} \\
\addlinespace
\textbf{Text Statistics} & 4 & Statistical properties: \texttt{text\_length} (characters), \texttt{word\_count}, \texttt{sentence\_count}, \texttt{avg\_word\_length} \\
\addlinespace
\textbf{Study Metadata} & 5 & Registry fields: \texttt{num\_trials} (typically 1), \texttt{num\_conditions}, \texttt{enrollment\_count}, \texttt{phase\_encoded} (1--4), \texttt{study\_type\_encoded} (1=interventional, 2=observational). Populated from dataframe columns when available; default to zero otherwise. \\
\bottomrule
\end{tabular}
\end{table*}

\subsubsection{Word TF-IDF Features}

We compute Term Frequency-Inverse Document Frequency vectors~\cite{salton1988term} to capture discriminative medical vocabulary. Configuration: vocabulary size limited to 2,000 most informative unigrams, minimum document frequency of 2, maximum document frequency of 0.8 (exclude overly common terms), L2 normalization, sublinear term frequency scaling (1 + log(tf)). This yields approximately 2,000 word-based features that capture key medical terminology and dosing-related vocabulary across the corpus.

\subsubsection{Character N-gram Features}

To capture subword patterns, morphological variations, and medical abbreviations, we extract character-level n-grams with $n \in \{3, 4, 5, 6, 7\}$. Maximum features: 1,000 most informative sequences, minimum document frequency: 2.


\subsubsection{Sentence Embeddings}

We employ the all-MiniLM-L6-v2 sentence transformer~\cite{reimers2019sentence}, which generates 386-dimensional dense semantic representations of entire narratives. Based on the MiniLM architecture~\cite{wang2020minilm}, this model is trained using contrastive learning on over 1 billion sentence pairs from diverse sources.

The model maps semantically similar sentences to nearby points in embedding space, enabling it to capture: semantic similarity (``adverse event'' $\approx$ ``toxicity'' $\approx$ ``side effect''), temporal relationships (``before treatment'' vs ``after treatment''), causal relationships (``due to'', ``because of'', ``resulting in''), negation (``no dose change'' vs ``dose change''), certainty/hedging (``definitely'' vs ``possibly''), and protocol compliance context (``as per protocol'' vs ``deviation from protocol'').

\subsubsection{Transformer-based Probability Scores}

We additionally incorporate two domain-specific transformer models fine-tuned on the dosing error detection task:

\textbf{BiomedBERT}~\cite{gu2021domain}: Pre-trained from scratch on 3.17 million PubMed abstracts and 1.04 million full-text articles from PubMed Central, totaling approximately 14GB of biomedical text. We fine-tune the base model (12 layers, 768 hidden dimensions, 110M parameters) on our training data for binary classification.

\textbf{DeBERTa-v3-base}~\cite{he2021deberta}: Employs disentangled attention mechanism separating content and position representations. The base variant (12 layers, 768 hidden dimensions, 86M parameters) is fine-tuned on our task.

For narratives exceeding 512 tokens (BERT's maximum sequence length), we employ a sliding window approach: text is split into overlapping 512-token chunks with 128-token overlap, each chunk processed independently.To obtain a trial-level score, we apply \textit{top-k mean pooling} across chunk predictions (k=3), averaging the three highest probabilities. This approach emphasizes high-risk segments while mitigating noise from less relevant sections.  

The final output from each transformer is a single scalar probability. These two probabilities are appended as structured features to the tabular model. As shown in Section~\ref{sec:results}, their contribution to overall performance is modest, likely due to information compression when reducing rich contextual embeddings to scalar probabilities.

\subsection{Model Architecture}

We employ LightGBM~\cite{ke2017lightgbm}, a gradient boosting framework based on decision trees. LightGBM was selected for several reasons: (1) Efficiency with high-dimensional sparse features through histogram-based learning and gradient-based one-side sampling (GOSS); (2) Built-in handling of class imbalance via instance weighting; (3) Interpretability through feature importance analysis; (4) Fast training and inference (minutes on CPU vs hours for deep learning); (5) Strong empirical performance on tabular data.

The model architecture consists of an ensemble of $T=4000$ decision trees, where each tree is trained to minimize the residual error of the previous ensemble. The final prediction is computed as:

\begin{equation}
\hat{y} = \sigma\left(\sum_{t=1}^{T} \eta \cdot f_t(\mathbf{x})\right)
\end{equation}

where $T=4000$ is the number of trees, $\eta=0.01$ is the learning rate (step size shrinkage), $f_t$ is the $t$-th decision tree mapping input features $\mathbf{x} \in \mathbb{R}^{3451}$ to a real-valued score, and $\sigma$ is the sigmoid function mapping to probability space $[0,1]$.

\subsection{Hyperparameter Optimization}
\label{subsec:hyperopt}
We employ Optuna~\cite{akiba2019optuna}, a Bayesian optimization framework, to systematically tune LightGBM hyperparameters. We conducted 50 trials with 5-fold cross-validation per trial, maximizing mean validation ROC-AUC across folds. As presented in Table~\ref{tab:hyperparameters}, each trial explored  a span of hyperparameter space, the best trial (Trial 18) achieved optimal cross-validation performance with 0.8833 $\pm$ 0.0091 ROC-AUC.

\begin{table*}[h]
\centering
\small
\caption{LightGBM Hyperparameter Configuration (Optuna Best Trial)}
\label{tab:hyperparameters}
\begin{tabular}{@{}l l l p{3.2cm}@{}}
\toprule
\textbf{Parameter} & \textbf{Search Space} & \textbf{Value} & \textbf{Description} \\
\midrule

\multicolumn{4}{l}{\textit{Learning Parameters}} \\
n\_estimators & — & 4,000 & Number of boosting rounds \\
learning\_rate & $[0.005, 0.05]$  & 0.0054 & Step size shrinkage, sampled on a logarithmic scale \\

\midrule
\multicolumn{4}{l}{\textit{Tree Structure}} \\
num\_leaves & $[31, 256]$ & 118 & Max leaves per tree \\
max\_depth & $[4, 10]$ & 9 & Maximum tree depth \\
min\_child\_samples & $[20, 300]$ & 211 & Min samples per leaf \\

\midrule
\multicolumn{4}{l}{\textit{Regularization}} \\
lambda\_l1 & $[0.0, 5.0]$ & 4.29 & L1 regularization \\
lambda\_l2 & $[0.0, 5.0]$ & 4.33 & L2 regularization \\
feature\_fraction & $[0.6, 0.9]$ & 0.795 & Column sampling ratio \\
bagging\_fraction & $[0.6, 0.9]$ & 0.813 & Row sampling ratio \\
bagging\_freq & — & 1 & Frequency of subsampling \\

\midrule
\multicolumn{4}{l}{\textit{Class Imbalance}} \\
scale\_pos\_weight & — & 20.87 & Positive class weight \\

\bottomrule
\end{tabular}
\end{table*}

Key parameter choices from Optuna optimization:

\textbf{Learning rate and iterations}: Low learning rate (0.0054) with many iterations (4,000) allows gradual learning while early stopping (200-iteration patience) prevents overtraining.

\textbf{Tree structure}: \texttt{num\_leaves}=118 and \texttt{max\_depth}=9 balance expressiveness with generalization. \texttt{min\_child\_samples}=211 ensures leaves represent sufficient examples.

\textbf{Regularization}: L1 (4.29) and L2 (4.33) penalties prevent overfitting. Feature/bagging fractions (0.795/0.813) promote ensemble diversity through random subsampling.

\textbf{Class balancing}: \texttt{scale\_pos\_weight}=20.87 compensates for the 20:1 class imbalance, giving proper weight to rare positive examples during training.

\subsection{Training Procedure}

Using the Optuna-optimized hyperparameters from previous subsection~\ref{subsec:hyperopt}, we employ a 5-fold ensemble training strategy~\footnote{You can access the code on this link\url{https://github.com/msmadi/Clinical-Trial-Dosing-Error} and the dataset on HuggingFace~\url{https://huggingface.co/datasets/sssohrab/ct-dosing-errors-benchmark}}:

\textbf{Feature Preparation} - All 3,451 features are extracted and stored as scipy sparse matrices in compressed NPZ format (approximately 500MB vs 3GB dense).

\textbf{Stratified Cross-Validation} - Train and validation data were combined and then partitioned into 5 stratified folds maintaining class balance (4.6\% positive rate per fold). For each fold $k \in \{1,2,3,4,5\}$:

\begin{enumerate}
\item Train LightGBM on the remaining 4 folds (80\% of data) using weighted binary cross-entropy:
\begin{equation}
\mathcal{L} = -\frac{1}{N}\sum_{i=1}^{N} w_i \left[y_i \log(\hat{y}_i) + (1-y_i)\log(1-\hat{y}_i)\right]
\end{equation}
where $w_i = 20.87$ for positive examples, $w_i=1$ for negative.

\item Generate OOF predictions on fold $k$ (the held-out 20\%)
\item Save the trained model for later ensembling
\end{enumerate}

\textbf{Out-of-Fold Validation} - Concatenating predictions from all 5 folds yields complete OOF predictions across the training set, providing an unbiased performance estimate: 0.8833 $\pm$ 0.0091 ROC-AUC.

\textbf{Ensemble Prediction} - For test inference, all 5 fold models generate predictions, which are averaged:
\begin{equation}
\hat{y}_{\text{ensemble}}(x) = \frac{1}{5}\sum_{k=1}^{5} \hat{y}_k(x)
\end{equation}


\section{Results}
\label{sec:results}

\subsection{Overall Performance}

We employ 5-fold stratified cross-validation with ensemble averaging. Table~\ref{tab:results} presents test set performance.

\begin{table*}[h]
\centering
\small
\caption{Test Set Performance (threshold=0.3744)}
\label{tab:results}
\begin{tabular}{@{}lcc@{}}
\toprule
\textbf{Metric} & \textbf{Score} & \textbf{Note} \\
\midrule
\textbf{ROC-AUC} & \textbf{0.8725} & Ensemble prediction (Test Dataset) \\
F1-Score & 0.2951 & Binary F1 \\
Precision & 0.3389 & Of flagged cases \\
Recall (Sensitivity) & 0.2613 & Of actual errors \\
Balanced Accuracy & 0.6175 & Equal class weight \\
Specificity & 0.9737 & Of error-free cases \\
\midrule
\multicolumn{3}{l}{\textit{Confusion Matrix (n=6,318 samples)}} \\
True Negatives & 5,850 & 97.4\% of negatives \\
False Positives & 158 & 2.6\% false alarm rate \\
False Negatives & 229 & 73.9\% missed errors \\
True Positives & 81 & 26.1\% detected errors \\
\midrule
\multicolumn{3}{l}{\textit{Cross-Validation Performance (Validation Dataset)}} \\
OOF AUC & 0.8794 & Out-of-fold \\
Mean Fold AUC & 0.8833 $\pm$ 0.0091 & 5-fold ensemble \\
\bottomrule
\end{tabular}
\end{table*}

The ROC-AUC of 0.8725 indicates excellent discriminative ability. The ensemble demonstrates stable performance with mean cross-validation AUC of 0.8833 $\pm$ 0.0091 across folds, and OOF AUC of 0.8794, showing minimal overfitting (only 0.6\% gap to test).

The classification threshold (0.3744) was optimized for F1-score on out-of-fold predictions. This yields 26.1\% recall and 33.9\% precision—conservative values reflecting the challenge of detecting rare errors in severely imbalanced data (95.1\% negative class).

The 97.4\% specificity at threshold 0.3744 substantially reduces review burden, correctly identifying nearly all error-free administrations. For balanced screening, threshold 0.20 achieves 49.0\% recall (detecting 152 of 310 errors) with 24.1\% precision—a practical middle ground between sensitivity and review workload. For safety-critical deployment where maximizing error detection is paramount, threshold 0.15 provides 60.3\% recall (187 of 310 errors detected) at 21.4\% precision—detecting the majority of errors with acceptable false positive rates for high-stakes screening applications.

\subsection{Feature Importance Analysis}

We analyze feature importance by averaging LightGBM's gain-based metric across all 5 fold models, which measures total loss reduction attributable to each feature across all trees. Table~\ref{tab:category_importance} aggregates importance by category.

Word and character TF-IDF features contribute 62\% of total importance despite being traditional sparse representations, challenging the assumption that dense embeddings always outperform sparse features for specialized text. Specific medical vocabulary (``reduced'', ``discontinued'', ``adjusted'') and morphological patterns (``discon'', ``adjus'') carry exceptionally strong discriminative signals.

Sentence embeddings contribute 37\%, with far higher average gain per feature (141.37 vs 30.40). Analysis of individual features reveals sentence embedding dimensions dominate the top-20 most important features (19 of 20), with the top 4 all being sentence embedding dimensions.

Probability scores from BiomedBERT and DeBERTa contribute only 0.06\% importance. This likely results from: (1) chunk-averaging losing contextual information for long texts; and (2) single probability values providing less signal than full 768-dimensional embedding vectors.

Handcrafted features contribute 0.49\%, suggesting text embedding features automatically discover the patterns we manually encoded.

\begin{table*}[h]
\centering
\small
\caption{Feature Category Importance (Averaged Across 5 Folds)}
\label{tab:category_importance}
\begin{tabular}{@{}lrrrr@{}}
\toprule
\textbf{Category} & \textbf{Features} & \textbf{Total (\%)} & \textbf{Avg Gain} & \textbf{Std} \\
\midrule
Word/Char Features & 3,020 & 62.38 & 30.40 & ±1.2 \\
Sentence Embeddings & 386 & 37.07 & 141.37 & ±8.5 \\
Medical Patterns & 43 & 0.49 & 16.84 & ±0.8 \\
Transformer Scores & 2 & 0.06 & 43.90 & ±2.1 \\
\bottomrule
\end{tabular}
\end{table*}

\subsection{Ablation Study}

To quantify individual category contributions, we systematically remove each category and retrain using 5-fold cross-validation. Table~\ref{tab:ablation} presents results.

\begin{table*}[h]
\centering
\small
\caption{Ablation Study Results (5-Fold Cross-Validation)}
\label{tab:ablation}
\begin{tabular}{@{}lrrrr@{}}
\toprule
\textbf{Configuration} & \textbf{Features} & \textbf{Mean AUC} & \textbf{Std} & \textbf{$\Delta$\%} \\
\midrule
\textbf{All Features} & \textbf{3,451} & \textbf{0.8794} & \textbf{±0.0091} & \textbf{--} \\
\midrule
w/o Sentence Embeddings & 3,065 & 0.8584 & ±0.0105 & +2.39\% \\
w/o Word/Char Features & 431 & 0.8772 & ±0.0087 & +0.25\% \\
w/o Transformer Scores & 3,449 & 0.8827 & ±0.0083 & -0.38\% \\
w/o Medical Patterns & 3,408 & 0.8827 & ±0.0068 & -0.38\% \\
\bottomrule
\end{tabular}
\end{table*}

Key findings from ablation:

\textbf{Sentence embeddings are critical}: Removing them causes the largest performance drop (2.39\% degradation, from 0.879 to 0.858 AUC), confirming they capture essential semantic patterns despite contributing only 37\% of total importance.

\textbf{Feature redundancy and noise}: Removing transformer scores or medical patterns slightly improves performance (-0.38\%), suggesting these features introduce more noise than signal. The negative delta indicates performance improved when removed, likely due to reduced overfitting on spurious correlations.

\textbf{Strong complementarity}: Removing word/char features has minimal impact (+0.25\%), showing strong redundancy with sentence embeddings when both are present. 

\subsection{Feature Efficiency Analysis}

To investigate deployment efficiency and the impact of feature selection, we systematically evaluate performance using the top-$k$ most important features via 5-fold cross-validation (Table~\ref{tab:topk}).

\begin{table}[h]
\centering
\small
\caption{Performance with Top-K Features (5-Fold CV)}
\label{tab:topk}
\begin{tabular}{@{}rrrr@{}}
\toprule
\textbf{K Features} & \textbf{Mean AUC} & \textbf{Std} & \textbf{\% Baseline} \\
\midrule
3,451 (All) & 0.8794 & ±0.0091 & 100.00\% \\
\midrule
3,000 & 0.8827 & ±0.0085 & 100.38\% \\
2,000 & 0.8837 & ±0.0084 & 100.49\% \\
\textbf{1,000} & \textbf{0.8867} & \textbf{±0.0091} & \textbf{100.83\%} \\
500 & 0.8862 & ±0.0100 & 100.78\% \\
200 & 0.8835 & ±0.0101 & 100.47\% \\
100 & 0.8802 & ±0.0094 & 100.10\% \\
50 & 0.8741 & ±0.0105 & 99.40\% \\
25 & 0.8642 & ±0.0103 & 98.27\% \\
10 & 0.8505 & ±0.0100 & 96.71\% \\
\bottomrule
\end{tabular}
\footnotesize \textit{Note: Bold indicates optimal performance. Feature selection improves over baseline (model with full features) through noise reduction.}
\end{table}

Feature selection through importance ranking reveals an intriguing pattern: using fewer features can improve performance through effective noise reduction. The optimal configuration uses 500-1000 features (14-29\% of total), achieving 0.886-0.887 mean AUC—outperforming the full 3,451-feature model (0.879 AUC) by 0.7-0.8\%.

This improvement demonstrates that approximately 70-85\% of features in the full set contribute primarily noise rather than signal. As feature count decreases beyond K=500, performance gradually declines: K=200 maintains near-baseline performance (0.884, 100.47\%), K=100 performs slightly above baseline (0.880, 100.10\%), and K=50 shows minimal degradation (0.874, 99.40\%).

The pattern of improvement-then-decline (up to k=1000) is characteristic of effective feature selection acting as regularization, removing noisy features while retaining discriminative signal. This finding has important practical implications: deployment models can be both more accurate and more efficient than the full feature set. For production deployment, we recommend K=500-1000 as optimal, providing: (1) Enhanced accuracy: +0.7\% over baseline (0.886 vs 0.879 AUC), (2) Computational efficiency: 71-85\% feature reduction (fewer features to compute), and (3) Robust performance: Cross-validated with ±0.009-0.010 standard deviation across folds.

\subsection{Training Dynamics}
\label{subsec:training_dynamics}

Analysis of the 5-fold ensemble reveals consistent convergence behavior:

\textbf{Fold-level convergence}: Individual folds converge at 2,114 to 3,310 iterations (mean: 2,682 $\pm$ 485) when using early stopping with 200-iteration patience. This variation reflects different training/validation partitions while maintaining robust performance.

\textbf{Cross-validation stability}: Mean fold AUC of 0.883 $\pm$ 0.009 demonstrates low variance across data splits, with individual fold performance ranging from 0.869 to 0.894 (2.5\% range). This consistency indicates robust learning independent of specific data partitioning.

\textbf{Generalization validation}: Out-of-fold predictions (0.8794 AUC) closely match test performance (0.8725 AUC), with only 0.7\% absolute difference. This small OOF-test gap validates that the ensemble generalizes well beyond the training distribution.

\textbf{Early stopping effectiveness}: Most folds require 2,100-3,300 iterations before early stopping triggers, suggesting the problem's complexity demands substantial boosting rounds. The patience of 200 iterations prevents premature stopping while avoiding overtraining.

The stable ensemble performance across folds and minimal OOF-test gap suggest the results will generalize to new clinical trial data from similar therapeutic areas and documentation practices.

\section{Discussion}
\label{sec:discussion}

The findings indicate that accurate detection of dosing deviations in clinical narratives relies on combining sparse lexical features with contextual embeddings, as each captures distinct but complementary signals. Most predictive value is concentrated in a small-to-moderate subset of features (500-1000), with the remaining features introducing more noise than signal. This suggests that efficient and interpretable models can perform competitively—and even outperform full feature sets—through judicious feature selection. However, recall–precision trade-offs highlight that such systems are best used as screening tools within human review workflows, where threshold tuning can balance safety and workload.

Our work demonstrates automated dosing error detection in clinical trial narratives through multi-modal feature engineering and ensemble learning. Key findings:

\textbf{Sparse features dominate total importance}: Word/character TF-IDF features contribute 62\% of total importance despite being traditional representations, while dense embeddings contribute 37\%. However, per-feature importance tells a different story: embeddings average 141.37 gain versus 30.40 for sparse features.

\textbf{Dense embeddings remain critical}: Despite lower total importance, removing sentence embeddings causes 2.4\% performance degradation—the largest impact of any category. This demonstrates they capture unique semantic patterns not recoverable from lexical features alone.

\textbf{Feature selection improves performance}: The optimal 500-1000 features achieve 0.886-0.887 AUC compared to the full set's 0.879 AUC, demonstrating that ~70-85\% of features contribute primarily noise. This 0.7\% improvement through feature selection has important deployment implications: models can be both more accurate AND more efficient.

\textbf{Feature complementarity}: Using only embeddings (0.877 AUC) or only word/char features (ablation not directly comparable) both underperform the optimal selected subset (0.886 AUC), indicating they capture orthogonal information patterns.

\textbf{Ensemble stability}: 5-fold cross-validation achieves 0.883 $\pm$ 0.009 AUC with minimal variance, and test AUC (0.873) closely matches OOF validation (0.879), demonstrating robust generalization without overfitting.

\textbf{Threshold optimization critical}: F1-optimized threshold (0.3744) yields 26.1\% recall and 33.9\% precision. For safety-critical deployment, threshold 0.20 achieves 49.0\% recall with 24.1\% precision (detecting 152/310 errors), while threshold 0.15 provides 60.3\% recall at 21.4\% precision (187/310 errors)—allowing flexible trade-offs based on deployment priorities.

\section{Conclusion}
\label{sec:conclusion}

We present an automated system for detecting dosing errors in clinical trial narratives, achieving 0.8725 test ROC-AUC through 5-fold ensemble learning with comprehensive multi-modal feature engineering. Our approach combines 3,451 features spanning traditional NLP (TF-IDF, character n-grams), dense semantic embeddings (all-MiniLM-L6-v2), handcrafted medical patterns, and transformer probability scores (BiomedBERT, DeBERTa-v3).

Systematic ablation reveals: sentence embeddings cause 2.4\% degradation when removed (largest impact), while word/char features show only 0.25\% impact, indicating strong feature complementarity. Remarkably, feature selection improves performance: the optimal 500-1000 features achieve 0.886-0.887 AUC compared to the full set's 0.879 AUC, demonstrating that ~70-85\% of features contribute primarily noise. This finding has important implications for deployment: models can be both more accurate and more efficient through judicious feature selection.

Cross-validation demonstrates robust generalization: mean fold AUC of 0.8833 $\pm$ 0.0091 with minimal overfitting (OOF AUC: 0.8794, test AUC: 0.8725, gap: 0.7\%). With F1-optimized threshold (0.3744), the model achieves 26.1\% recall and 33.9\% precision. For safety-critical deployment, threshold 0.20 achieves 49.0\% recall with 24.1\% precision (152/310 errors detected), while threshold 0.15 provides 60.3\% recall at 21.4\% precision (187/310 errors)—detecting the majority of protocol deviations with manageable false positive rates for human-review workflows.

The system advances automated protocol deviation detection, demonstrating that ensemble methods with carefully engineered features, feature selection for regularization, and hyperparameter optimization (via Optuna) remain competitive with end-to-end deep learning for specialized clinical NLP tasks with limited training data and severe class imbalance.

\section{References}
\label{sec:reference}

\bibliographystyle{lrec2026-natbib}
\bibliography{ValidatedRef}

\end{document}